\pdfoutput=1
\documentclass[11pt,a4paper]{article}
\usepackage{acl}
\usepackage{times}
\usepackage{latexsym}

\usepackage{microtype}

\usepackage{hyperref}
\PassOptionsToPackage{linktocpage}{hyperref}

\usepackage{url}
\usepackage{booktabs}
\usepackage{amsfonts}
\usepackage{amsmath,amsthm}
\usepackage{graphicx}
\usepackage{mathrsfs}
\usepackage{enumitem}
\usepackage{soul}

\usepackage{breakurl}

\DeclareMathOperator*{\argmax}{arg\,max}

\aclfinalcopy % Uncomment this line for the final submission
\title{Learning to Match Mathematical Statements with Proofs}

\author{Maximin Coavoux\Thanks{ Work mostly done at the University of Edinburgh.}\\
  Univ. Grenoble Alpes, CNRS\\Grenoble INP, LIG\\
  \texttt{maximin.coavoux}\\\texttt{@univ-grenoble-alpes.fr} \\\And
  Shay B. Cohen \\
  ILCC, School of Informatics\\ University of Edinburgh\\
  \texttt{scohen@inf.ed.ac.uk} \\}

\date{}

\begin{document}
\maketitle
\begin{abstract}
We introduce a novel task consisting 
in assigning a proof to a given mathematical statement.
The task is designed to improve the processing
of research-level mathematical texts.
Applying Natural Language Processing (NLP) tools to research level
mathematical articles is both challenging, since it is a highly
specialized domain which mixes natural language and mathematical formulae.
It is also an important requirement for developing tools
for mathematical information retrieval and computer-assisted theorem proving \cite{2014arXiv1404.1905M}.
We release a dataset for the task,
consisting of over 180k
statement-proof pairs extracted from
mathematical research articles.
We carry out preliminary experiments to assess the
difficulty of the task.
We first experiment with two bag-of-words baselines.
We show that considering the assignment problem
globally and using weighted bipartite matching
algorithms helps a lot in tackling the task.
Finally, we introduce a self-attention-based model
that can be trained either locally or globally
and outperforms baselines by a wide margin.
\end{abstract}

\section{Introduction}

Research-level mathematical discourse is a challenging domain for
Natural Language Processing (NLP).
Indeed, mathematical articles switch frequently between natural
language and mathematical formulae.
A semantic analysis of mathematical text needs to solve
relationships (e.g.\ coreference) between mathematical symbols
and concepts.
Moreover, mathematical writing follows a lot of conventions,
such as variable naming or typography, that are implicit,
and may differ from a subfield to another.

However, mathematical research can benefit from NLP \cite{2014arXiv1404.1905M},
in particular as concerns bibliographical research:
researchers need tools to find work relevant for their research.
Indeed, prior NLP work on mathematical research articles
focused on Mathematical Information Retrieval (MIR)
and related tools or data \cite{DBLP:conf/ntcir/ZanibbiAKOTD16,C16-1221,P15-2055}.

\begin{figure}
  \fbox{
  \parbox{0.93\columnwidth}{
  \textbf{Theorem 1.3.} \textit{Suppose that $|\mathrm{Sing}(S)|<(2r-1)r$. Then $X$ is factorial.}\\
  \textit{Proof.} The subset $\mathrm{Sing}(S) \subset \mathbb{P}^3$ is a set-theoretic intersection 
  of surfaces of degree $2r-1$, which implies that $X$ is factorial by Theorem 1.1.
  }
}
  \caption{Example of a statement-proof pair.}
  \label{fig:pair-example}
\end{figure}

In this paper, we introduce a task
aimed at improving the processing of research-level
mathematical articles and make a step towards the
modeling of mathematical reasoning.
Given a collection of mathematical statements
and a collection of mathematical proofs of the same
size, the task consists in finding and assigning
a proof to each mathematical statement.
We construct and release a dataset for the task, by collecting
over 180k statement-proof pairs from mathematical research articles
(an example is given in Figure~\ref{fig:pair-example}).

There are multiple motivations for the design of the task.
We believe it may help MIR by serving as a proxy for 
the search for the existence of a mathematical
result, or for theorems and proofs related to one another (e.g.\ using the same
proof technique), an important search tool for any digital mathematical library 
\cite{2014arXiv1404.1905M}.
Learning to match statements and proofs would also benefit computer-assisted 
theorem proving, as it is akin to tasks such as premise selection, 
also recently addressed with NLP methods \cite{DBLP:journals/corr/abs-1905-07961}.
More generally, finding supporting information for or against a given statement,
is integral to tasks such as question answering
or fact-checking \cite{vlachos-riedel:2014:W14-25}.
Our mathematical statement-proof assignment task can be thought
of as the transposition of such problem to the very specific
domain of mathematical research articles.

We provide preliminary results on our proposed task with (i) two bag-of-words baselines
(ii) a neural model based on a self-attentive encoder  and a bilinear similarity function.
Though the neural model outperforms the baselines when
using \textit{local decoding}, i.e.\ assigning
the best-scoring proof to each statement,
we found that it performs even better with \textit{global decoding},
i.e.\ finding the best bipartite matching between the sets of statements and proofs.
Therefore we also design a global training procedure with
a structured max-margin objective.
Such an architecture may have applications to other
NLP problems that can be cast as maximum bipartite matching problems,
which is the case, for example, for some alignment problems \cite{H05-1010,pado-lapata:2006:COLACL}.

In summary, our contributions are three-fold:
\begin{itemize}[noitemsep]
    \item The definition of a mathematical statement-proof matching task;
    \item The construction and release of a corresponding dataset;
    \item A self-attention-based model for maximum weighted
    bipartite matching problems, that can be trained either locally or globally.
\end{itemize}

\section{Related Work}

\paragraph{Processing mathematical articles}
Most NLP work on mathematical discourse focuses on improving
Mathematical Information Retrieval \cite[MIR]{DBLP:conf/ntcir/ZanibbiAKOTD16} by 
establishing connections between mathematical formulae and natural language text
in order to improve the representation of formulae.

The interpretation of variables is highly dependent on the context.
For example, the symbol $E$ could denote an expectation in a statistics article,
or the energy in a physics article.
Some studies use the surrounding context of a formula
to assign a definition or a type to the whole formula, or to specific variables.
\newcite{W10-3910} focus on identifying coreferences between mathematical
formulae and mathematical concepts in Wikipedia articles.
\newcite{Kristianto12extractingdefinitions} extract definitions of mathematical expressions.
\newcite{zbMATH05679807}, \newcite{wolska2011} and \newcite{Schubotz:2016:SIM:2911451.2911503}
disambiguate mathematical identifiers, such as variables, using the surrounding textual context.
\newcite{N18-1028} infer the type of a variable in a formula from the textual context of the formula.

Another line of work focused on identifying specialized terms or concepts
to improve MIR \cite{P15-2055,C16-1221}.

Some work adapts standard NLP tools to the specificity of mathematical discourse,
e.g.\ POS taggers \cite{DBLP:journals/corr/SchonebergS14},
with the objective of using linguistic features to improve the search for definitions 
of mathematical expressions \cite{DBLP:journals/corr/PagaelS14}.

\paragraph{Maximum bipartite matching in NLP}
Global models for maximum weighted bipartite matching problems
have been explored in NLP for the task of word alignments,
a traditional component of machine translation systems
\cite{C04-1032,H05-1010,S12-1085,Y16-2012},
or for assigning arguments to predicates \cite{Q13-1018}.
In particular, \newcite{H05-1010} introduced a discriminative
global model with a max-margin objective.

In these articles, the bipartite graph is usually formed by two
sentences. In contrast, we predict matchings on graphs that are an order
of magnitude larger and
each node in our bipartite graph is a complete
text (a statement or a proof), i.e.\ a highly structured object,
from which we learn fixed size vector representations.

\section{Task Description}

Given a collection of mathematical statements $\{s_i\}_{i \leq N}$, and a separate equal-size collection of mathematical proofs $\{p_i\}_{i \leq N}$, we are interested in the problem of assigning a proof to each statement.

\paragraph{Evaluation} We use two evaluation metrics.
Assumming that a system predicts a ranking of proofs,
instead of providing only a single proof, we evaluate
its output with the Mean Reciprocal Rank (MRR) measure:
\[ \text{MRR}(\{\hat r_i \}_{i \in \{1, \dots N\}}) = \frac{1}{N} \sum_{i=1}^N \frac{1}{\hat r_i}, \]
where $N$ is the number of examples and $\hat r_i$
is the rank of the gold proof for statement number $i$,
as predicted by the system.

As a second evaluation metric, we use a simple accuracy,
i.e.\ the proportion of statements whose first-ranked proof
is correct.

By construction (see Section~\ref{sec:data}), it is possible though unlikely that the same mathematical statement occurs several times in the dataset.
It is all the more unlikely that several occurrences have exactly the same formulation and use the same variable names. Therefore, we consider a match to be correct if and only if it is associated with its original proof.

\paragraph{Task variation}
We propose three variations of the task, depending on the input of the system:
\begin{enumerate}[noitemsep]
    \item Natural language text and mathematical formulae;
    \item Natural language text only; 
    \item Mathematical formulae only.
\end{enumerate}
The comparison of these settings is meant to provide insight
into which type of information is crucial to the task.

\section{Dataset Construction}
\label{sec:data}

This section describes the construction of a dataset\footnote{\url{https://gitlab.com/mcoavoux/statement_proof_matching}}
of statement-proof pairs (see Figure~\ref{fig:pair-example}).

\paragraph{Source corpus} 
We use the MREC
corpus\footnote{\url{https://mir.fi.muni.cz/MREC/}, version 2011.4.439.} \cite{dml:702604}
as a source.
The MREC corpus contains around 450k articles from ArxMLiV \cite{stamerjohanns2010transforming},
an on-going project aiming at converting the arXiv\footnote{\url{https://arxiv.org/}}
repository from \LaTeX~to XML, a format more suited to machine processing.
In this collection, mathematical formulae are represented in the
MathML\footnote{\url{https://www.w3.org/Math/}} format,
a markup language.

\paragraph{Statement-proof identification}

For each XML document (corresponding to a single arXiv article),
we extract pairs of consecutive \texttt{<div>} tags such that:
%\begin{itemize}[noitemsep]
    %\item 
    (i) the \texttt{class} attribute of the first \texttt{div} node
    contains the string \texttt{"theorem"};
    %\item
    (ii) the \texttt{class} attribute of the second \texttt{div} node
    is the string \texttt{"proof"}.
%\end{itemize}
Documents that do not contain such pairs of tags are discarded,
as well as documents that are not written in English
(representing 143 articles in French,
11 in Russian, 5 in German, 2 in Portuguese and 1 in Ukrainian),
as identified by the \texttt{polyglot} Python package.\footnote{\url{www.github.com/aboSamoor/polyglot/}}

In the remaining collection of pairs of statements and proofs,
we filter out pairs for which either the statement or the proof is too short.\footnote{We used a minimum length
of 20 tokens for both statements and proofs, based on a manual inspection of the shortest examples. We also exclude proofs and statements longer than 500 tokens.}
Indeed, the short texts were often empty (only consisting of a title, e.g.\ ``5.26 Lemma.''),
which we attribute to the noise inherent to the conversion to XML,
or not self-contained.
In particular, we identified several prototypical cases:
\begin{itemize}[noitemsep]
 \item Omitted (or easy) proofs contain usually a single word
 (`omitted', `straightforward',
 %`immediate',
 `well-known',
 %`easy',
 `trivial',
 %`obvious', `clear',
 `evident'),
 but are sometimes more verbose (`This is obvious and will be left to the readers').
% `This is tedious but straightforward.').
 \item Proofs that consist of a single reference to
 \begin{itemize}[noitemsep]
 \item An appendix (`See Appendix A');
 \item Another theorem (`This follows immediately from Proposition 4.4 (ii).');
 \item The proof method of another theorem (`Similar to proof of Lemma 6.1')
 \item Another article (`See [BK3, Theorem 4.8].');
 \item Another part of the article
 (`The proof will appear elsewhere.', `See above.', `Will be given in section 5.').
 \end{itemize}
\end{itemize}
Filtering on the number of tokens also exclude self-contained short proofs,
such as `Take $Q\prime = p h_i - p_i$.'
However, such proofs were very infrequent on manual inspection
of the discarded pairs (2 in a manually inspected random sample of~100 discarded proofs).

\paragraph{Preprocessing: linearizing equations}
Mathematical formulae in the XML documents are enclosed
in a \texttt{<math>} markup tag, that materializes the switch to the MathML format,
and whose internal structure represents the formula as an XML tree.
As a preprocessing step, we linearize each formula to a raw sequence of strings.

In MathML, an equation can be encoded in a content-based (semantic) way or in a presentational way,
using different sets of markup tags.
We first convert all MathML trees to presentational MathML
using the XSL stylesheet from the Content MathML Polyfill repository.\footnote{\url{https://github.com/fred-wang/webextension-content-mathml-polyfill}}
Then we perform a depth-first search on each tree rooted in a \texttt{<math>} tag
to extract the text content of the whole tree.

During this preprocessing, we tested several processing choices:
\begin{itemize}[noitemsep]
 \item \textbf{Font information}. In mathematical discourses, fonts play an important role.
  Their semantics depend on conventions shared by researchers.
  %For example, 
  If both $x$ and $\mathbf x$ appear in the same article,
  they are most likely to represent different mathematical objects, e.g.\ a scalar and a vector.
  %When a font is specified for a mathematical symbol, we prefix the font name to the corresponding symbol.
% \item \textbf{Silent markup tags}. A number of tags in MathML are not rendered by textual content
% but (i) specify the position of an element, e.g.\ an exponent in superscript position,
% or (ii) represent mathematical expressions that have scopes over complex expressions,
% e.g.\ fractions or roots.
% We materialize each of these tags with a unique pseudo-token.
  Therefore, we use distinct symbols for tokens that are in distinct fonts.
  \item \textbf{Math-English ambiguity}. Some symbols can be used
  both in natural language text and in formulae. 
  For example, `a' can be a determiner in English, or a variable name in a formula.
  To avoid increasing ambiguity when linearizing formula, we type each symbol (as math or text)
  to make the mathematical vocabulary completely disjoint from the text vocabulary.
\end{itemize}
Both these preprocessing steps had a beneficial effect on the baselines in preliminary experiments.

%\paragraph{Obfuscating identifiers} 
%In mathematical discourse, the choice of variable names
%is both conventionalized and arbitrary.
%If all variable names are changed consistently in a mathematical statement,
%its truth value remains the same.

%We instantiate a second version of the dataset,
%where all mathematical identifiers are replaced by placeholders.
%In order to enforce consistency, we made sure if an identifier $x$
%appears both in a statement and in the corresponding proof,
%then, it is replaced by the same placeholder in both texts.

\paragraph{Statistics}
We report in Table~\ref{tab:statistics} some statistics about the dataset.
The extracted articles were from a diverse set of mathematical subdomains,
and connected domains, such as computer science (746 articles from 30 subcategories)
and mathematical physics (2562 from 31 subcategories).
There are in average 6.6 statement-proof pairs per article.

\begin{table}[t!]
    \resizebox{\columnwidth}{!}{
        \begin{tabular}{lrr}
            \toprule
            Number of documents in the MREC corpus          &&  439,423 \\
            Extracted documents with statement-proof pairs  &&   27,841 \\
            Total number of statement-proof pairs           &&  184,094 \\
            Number of (primary) categories            &&  (120) 135\\
            Average number of categories per article  &&    1.7\\
            \midrule
            \midrule
            Most represented primary categories & Num.\ articles & Num.\ pairs \\
            \midrule
math.AG Algebraic Geometry 		&	2848 &	22029 \\
math.DG Differential Geometry  &	2030 &	12440 \\
math.CO Combinatorics 			&	1705 &	10548 \\
math.GT Geometric Topology		&	1539 &	9234 \\
math.NT Number theory          &	1454 &	9521 \\
math.PR Probability				&	1422 &	7660 \\
math.AP Analysis of Partial Differential Equations &	1386 &	6981 \\
math-ph Mathematical Physics    &	1249 &	6491 \\
math.FA Functional Analysis    &	1143 &	8011 \\
math.GR Group Theory &	970 &	7806 \\
math.DS Dynamical System &	961 &	6424 \\
math.QA	Quantum Algebra & 944	& 8074 \\
math.OA	Operator Algebras  & 923	&     8050 \\
            \bottomrule
        \end{tabular}
    }
    \caption{Statistics about the dataset and categories of mathematical articles.}
    \label{tab:statistics}
\end{table}

\begin{table}
    \resizebox{\columnwidth}{!}{
        \begin{tabular}{lrrr}
            \toprule
            Statements   & Min   & Max & Mean$\pm$SD \\
            \midrule
            Text+math  &    20  &  500    & 80$\pm$57 \\
            Text only  &    1   &  398    & 30$\pm$20 \\
            Math only  &    0   &  470    & 58$\pm$20 \\
            Math proportion & 0$\%$ & 99.5$\%$ & 58$\%\pm$20\\
            \midrule
            Proofs     & \\
            \midrule
            Text+math  & 20 & 500 & 210$\pm$ 127 \\
            Text only  & 1 & 467  & 81  $\pm$ 56 \\
            Math only  & 0 & 495  & 129 $\pm$ 96 \\
            Math proportion & 0$\%$ & 99.6$\%$  & 56$\%\pm$ 21 \\
            \bottomrule
        \end{tabular}
    }
    \caption{Number of tokens in the dataset. We report for statements
    and proofs the minimum, maximum and average number of tokens
    broken down by type (`math' for tokens extracted from
    formulae and `text' for the others).
    A value of 0 for, e.g.\ the `math only' row, means that the statement or proof does not contain mathematical symbols or formulae.}
    \label{tab:token-stats}
\end{table}

We report statistics about the size of statements
and proofs in number of tokens in Table~\ref{tab:token-stats}.
We report the number of tokens in formulae 
(math), in the text itself (text) and in both (text+math).
On average, proofs are much longer than statements.
Statements and proofs have approximately the
same proportion of text and math.
Overall, the variation in number of tokens across statements and
proofs is extremely high, as illustrated by the standard deviation (SD)
of all presented metrics.

\section{Self-Attentive Bilinear Similarity Model}

We propose a system based on a self-attentive encoder \cite{NIPS2017_7181}
that constructs fixed-size vector representations for statements and proofs,
and a similarity function that scores the relatedness
of a statement-proof pair.

\paragraph{Self-attentive encoder}
We encode each text with a token-level self-attentive encoder.
We first project a text to a sequence of token embeddings of dimension $w$.
Then we run $\ell$ self-attention layers \cite{NIPS2017_7181}, to obtain 
a contextualized embedding for each token.
Finally we construct a vector representation for the text with a max-pooling
layer over the contextualized embeddings of the last self-attention layer.

The hyperparameters of the encoder are the dimension of the token embeddings $w$,
the number of self-attentive layers $\ell$,
the dimension of the encoder $d$ (size of contextualized embeddings),
the number of heads for each self-attentive layer $h$
and the dimension of query and key vectors $d_k$.

\paragraph{Trainable bilinear similarity function}
Given the encoded representations of a statement $\mathbf s = \text{enc}(s)$
and a proof $\mathbf p = \text{enc}(p)$, we compute an association score
with the following bilinear form:
\begin{equation*}
    \text{score}(\mathbf s, \mathbf p) = \mathbf s^{\top} \cdot \mathbf W \cdot \mathbf p + b,
\end{equation*}
where $\mathbf W$ and $b$ are parameters that are learned together with
the self-attentive module parameters.

\paragraph{Local decoding}
For a collection of $n$ statements and proofs, we first score
all possible pairs ($s, p$), and construct a matrix $M = (m_{ij}) \in \mathbb{R}^{n\times n}$,
with
\begin{equation*}
m_{ij} = \text{score}(\mathbf s^{(i)}, \mathbf p^{(j)}),
\end{equation*}
where $\mathbf s^{(i)}$ and  $\mathbf p^{(j)}$ are
the encoded representations of, respectively, 
the $i^{th}$ statement and the $j^{th}$ proof.
Then we can straightforwardly sort each row by decreasing order and assign
the proof ranking to the corresponding statement.
The best ranking proof $\hat p$ for statement $i$ satisfies:
\[ \hat p_i = \argmax_{j} \; m_{ij}. \]
We call this decoding method `local', since it does not take
into account dependencies between assignments.
In particular, several statements may have the same highest-ranking proof.

\paragraph{Global decoding}
The local decoding method overlooks a crucial piece of information:
a proof should correspond to a single statement.
In a worst-case situation, a small number of proofs may score
high with most statements and be systematically assigned as
highest-ranking proof by the local decoding method.

During preliminary experiments, we analysed the output
of our system with local decoding on the development set,
focusing on the distribution of the single highest-ranking proof for each statement.
It turned out that around $23\%$ of proofs were assigned to at least
two different statements, whereas more than $40\%$ of proofs
were assigned to no statement (Table~\ref{tab:cumulative}).

\begin{table}
    \begin{center}
    \begin{tabular}{lrr}
        %\multicolumn{2}{c}{S}
        \toprule
        Statements & Proofs & $\%$\\
        \midrule
        $\geq 20$  & 7 & 0.0 \\
        $\geq 10$  & 80 & 0.2 \\
        $\geq 5$  & 1027 & 1.9 \\
        $\geq 2$  & 11949 & 22.6 \\
        $=1$  & 19531 & 37.0 \\
        $<1$  & 21275 & 40.3 \\
        \bottomrule
    \end{tabular}
    \end{center}
    \caption{Cumulative distribution of proofs in the development set,
    by number of statements to which they are assigned
    with the local decoding method.}
    \label{tab:cumulative}
\end{table}

We propose a second decoding method based on a global constraint on the output:
a proof can be assigned only to a single statement.
Intuitively, the constraint models the fact that 
if a proof is assigned by the system to a certain statement with high confidence,
we can rule it out as a candidate for other statements.
Under this constraint, the decoding problem reduces to
a classical maximum weighted bipartite matching problem,
or equivalently, a Linear Assignment Problem (LAP).
In more realistic scenarios (e.g.\ if the input sets of statements and proofs do not have the same size), the method would require some adaptation.

Formally, we define an assignment $A$ as a boolean
matrix $A = (a_{ij})\in \{0,1\}^{n\times n}$ with the following constraints:
\begin{align*}
    \forall i \forall j, \sum_{j} a_{ij} = \sum_{i} a_{ij} = 1,
\end{align*}
i.e.\ each row and each column of $A$ contains a single non-zero coefficient.
The score of an assignment $A$ is the sum of scores of the chosen edges:
\begin{align*}
    \text{score}(A,M) = \sum_i \sum_j a_{ij} m_{ij}.
\end{align*}
Finally, global decoding consists in solving the following LAP:
\begin{align*}
    \hat A(M) = \argmax_{\substack{A \in \{ 0,1 \}^{n\times n} \\\text{s.t.\ } \forall i \forall j, \sum_{j} a_{ij} = \sum_{i} a_{ij} = 1}}  \text{score}(A, M).
\end{align*}

The LAP is solved in polynomial time by
the Hungarian algorithm \cite{Kuhn1955},
the LAP-Jonker-Volgenant algorithm \cite[LAP-JV;][]{Jonker1987},
or the push-relabel algorithm \cite{Goldberg1995}.
These methods have a $\mathcal{O}(n^3)$ time complexity where $n$
is the number of pairs, and $\mathcal{O}(n^2)$ memory complexity.
This is too expensive in our case, due to the size
of our datasets (more than 18,000 pairs in the development set).

To remedy this limitation, when we perform decoding on a large set,
we only consider the $k$ best-scoring
proofs (i.e.\ outgoing edges in the bipartite graph)
for each statement, which makes the number of edges linear
in the number of pairs $n$ (considering $k$ fixed).
Moreover, we use a modification of the LAP-JV algorithm specifically
designed for sparse matrices \cite[LAP-MOD;][]{VOLGENANT1996917}.

\section{Local and Global Training}

We propose two training methods for the similarity
model introduced above:
a local training method that
only considers statements in isolation (Section~\ref{sec:local})
and a global model trained to predict a bipartite matching
(Section~\ref{sec:global}), with a hybrid global and local objective.

\subsection{Local Training}
\label{sec:local}

We would like to
train our model to assign a high similarity to the gold
statement-proof pair, and a low similarity to all other
statment-proof pairs.
This corresponds to the following objective, for a single
statement $s$ and its gold proof $p$:
\begin{align*}
    \mathcal{L}_{\textsc{loc}}(s, p, P; \boldsymbol \theta) &= - \log \mathbb P(p | s; \boldsymbol \theta) \\
                      &= - \log \left( 
\dfrac{e^{\text{score}(\mathbf s, \mathbf p)}}{\sum\limits_{p' \in P} e^{\text{score}(\mathbf s, \mathbf p')}}
\right),
\end{align*}
where $P$ is the set of proofs, and $\boldsymbol \theta$ are the parameters of the model.
Directly optimizing this loss function requires the computation
of $\mathbf p = \text{enc}(p)$ for every proof in the dataset,
for a single optimization step.
This is not realistic considering memory limitations,
the size of the train set,
and the fact that the self-attentive encoder is
the most computationally expensive part of the network.

Instead, we sample minibatches of $b$ pairs and optimize the following
proxy loss for the sequence $S'=(s_1, \dots, s_b)$ of statements
and the sequence $P'=(p_1, \dots, p_b)$ of corresponding proofs:\footnote{We also
experimented with a Noise-Contrastive Estimation approach \cite{journals/jmlr/GutmannH12}.
However, it exhibited a much slower convergence rate.}
\begin{equation*}
    \mathcal{L'}_{\textsc{loc}}(S', P'; \boldsymbol \theta) = \sum_{i=1}^b \mathcal{L}_{\textsc{loc}}(s_i, p_i, P'; \boldsymbol \theta).
\end{equation*}
In practice, we sample uniformly and without replacement $b$ pairs from 
the training set at each stochastic step.

\subsection{Hybrid Local and Global Training}
\label{sec:global}

The local training method only considers statements in isolation.
Even though we expect a locally trained model to perform better
with global decoding, we hypothesize that a model that is trained
to predict the full structure (a bipartite matching)
will be even better.

For a collection of $n$ proofs and $n$ statements,
the size of the search space (i.e.\ the number of bipartite
matchings) is $n!$, since each matching corresponds
to a permutation of proofs.
As a result, the use of a globally normalized model
is impractical.
We turn to a max-margin model that does not
require normalization over the full search space.

We use the following max-margin objective, for a set $B$ of $n$ pairs corresponding to matrix $M$:
\begin{align*}
    \mathcal{L}_{\textsc{global}}(B;\boldsymbol \theta) = \max(0, &\Delta(\hat A, I) \\ & +  \text{score}(\hat A, M)\\& - \text{score}(I, M)),
\end{align*}
where $\boldsymbol \theta$ is the set of all parameters
$\hat A$ is the predicted assignment and $I$ is
the gold assignment, i.e.\ the identity matrix.
The structured cost 
\begin{equation*}
\Delta(\hat A, I) = \sum_{ij} \max(0, (\hat A - I)_{ij})
\end{equation*}
aims at enforcing a margin for each individual assignment.
In order to compute the loss during training, we perform
decoding on matrix $M'$, which directly
incorporates the cost of wrong assignments \cite{Taskar:2005:LSP:1102351.1102464}:
\begin{equation*}
M' = M + (\mathbf{1} - I).
\end{equation*}

The computation of this loss requires exact decoding for each optimization step.
Since exact decoding is only feasible for a small $n$,
and since we need to keep track of all intermediary
vectors to compute the backpropagation step,\footnote{In particular,
the computation graph needs to conserve all encoding
layers for the $2n$ texts involved.}
we perform each stochastic optimization step on a minibatch
of pairs of size $b$.
Since this global objective had a very slow convergence rate (see Section~\ref{sec:setup}),
in practice, we optimize a hybrid local-global objective: $\mathcal{L'}_{\textsc{loc}} + \mathcal{L}_{\textsc{glob}}$.

\section{Experiments}

Our experiments address several questions.
First, we assess the difficulty of the task and provide
preliminary results with baseline systems.
Secondly, we evaluate the performance
of our neural model in several settings:
global or local training, global or local decoding.
In particular, we are interested in assessing
whether global decoding improves accuracy when training
is only local, and how the more complex global training
method fares with respect to local training.
Finally, we are interested in the informativeness
of different types of input: text, mathematical formulae, or both.

We describe the experimental protocol (Section~\ref{sec:setup})
before discussing results (Section~\ref{sec:results}).

\subsection{Experimental setup}
\label{sec:setup}

\paragraph{Dataset} We use the dataset whose construction
is described in Section~\ref{sec:data}.
We shuffle the collection of statement-proof pairs before
performing a $80\%/10\%/10\%$ train-development-test split,
corresponding to 147276 pairs for the training sets and 18409 pairs
for the development and tests.
Due to the shuffling, pairs from a single article may
be distributed across the three sections.

\paragraph{Baselines}

We provide two baseline systems that rank proofs according to their similarity to the statement,
using classical similarity measures.
The first baseline computes cosine similarities between
TF-IDF representations of statements and proofs.
The second baseline uses Dice's similarity measure computed over
bag-of-word representations of statements and proofs:
\[ \mathrm{Dice}(s, p)= \dfrac{2|s \cap p|}{|s| + |p|}, \]
where $s$ and $p$ are the word multiset representations of, respectively,
a statement and a proof.

Both baselines are implemented using the scikit-learn
Python package \cite{scikit-learn} with default parameters.
We estimate the IDF metric on the training set only.

\paragraph{Neural model}
We implemented the neural network in Pytorch \cite{paszke2017automatic}.
Token embedding have $c=300$ dimensions,
we use $\ell=2$ self-attentive layers with 4~heads to obtain contextualized embeddings
of dimension $d=300$.
The query and key vectors have size $d_k=128$.

We trained each model on a single GPU using the Pytorch implementation
of the Averaged Stochastic Gradient Descent algorithm
\cite[ASGD][]{Polyak:1992:ASA:131092.131098}, with learning rate~$0.02$,
and an exponential learning rate scheduler (the learning rate is multiplied by $0.99$ after each epoch).

\paragraph{Hyperparameters}
For training a local model, we perform~400 epochs
over the whole training set,
assuming an epoch consists in $N/b$ stochastic steps
(where $N$ is the total number of training pairs
and $b$ is the number of pairs in each minibatches).
We evaluate the model's performance on the development
set every~20 epochs and select the best model 
among these intermediate models.
We use batches of size~$b=60$ based on preliminary experiments.

For global training, we perform~400 epochs (around~3 days with a single GPU) %\footnote{Training a model takes about 3 days with a GPU.}
and use the same model selection method as in the local training experiments.
We observed in initial experiments that training only with the global objective
required a very long time and had a very slow
convergence rate.
Therefore, we used the following global-local objective:
$\mathcal{L'}_{\textsc{loc}} + \mathcal{L}_{\textsc{glob}}$,
that we optimized by alternating one stochastic step for each loss.
We use batches of size~60 for both the local loss and the global loss.
Although the global model might benefit from larger batches,
60 was the maximum possible size given our memory resources.

\paragraph{Global decoding}
Recall that exact global decoding is only feasible for a small subset of pairs.
During global training, we chose a batch size small enough to perform exact decoding.
However, it is not feasible to perform exact decoding on the whole development and test corpora.
Therefore, we prune the search space by keeping only the 500-best
candidate proofs for each statement, and use the LAP-MOD algorithm designed for sparse matrices.
In practice, we used the implementations of the LAP-JV and LAP-MOD algorithms
from the \texttt{lap} Python package,\footnote{\url{https://github.com/gatagat/lap}}
for respectively exact decoding on minibatches during global training
and decoding on whole datasets during evaluation.

\begin{table}
\begin{center}
\resizebox{\columnwidth}{!}{
\begin{tabular}{llccc}
    \toprule
& & \multicolumn{2}{c}{Local decoding} & Global decoding \\
    Input & Method & MRR & Accuracy & Accuracy\\
    \midrule
    \multicolumn{5}{l}{Dev} \\ 
    \midrule
        Both & Dice & 16.6 & 12.7 & 25.2 \\ 
        Both & TF-IDF & \textbf{29.9} & \textbf{23.8} & \textbf{36.3} \\ 
        Text & Dice & 10.4 & 7.8 & 16.2 \\ 
        Text & TF-IDF & 27.9 & 22.7 & 26.3 \\ 
        Math & Dice & 13.3 & 10.0 & 10.4 \\ 
        Math & TF-IDF & 12.1 & 9.1 & 9.5 \\ 
    \midrule
    \multicolumn{5}{l}{Test} \\
    \midrule
        Both & Dice & 16.8 & 12.9 & 25.4 \\ 
        Both & TF-IDF & 31.2 & 25.0 & 35.6 \\ 
        Text & Dice & 10.7 & 8.0 & 17.3 \\ 
        Text & TF-IDF & 27.8 & 22.4 & 26.4 \\ 
        Math & Dice & 13.6 & 10.2 & 11.1 \\ 
        Math & TF-IDF & 12.2 & 9.3 & 9.7 \\ 
    \bottomrule
\end{tabular}
}
\end{center}
\caption{Baseline results with the TF-IDF system and the word-overlap system (Dice),
with either global or local decoding. The input to the systems
are either only the textual parts, only the mathematical formulae, or both.}
\label{tab:baselines}
\end{table}

\subsection{Results}
\label{sec:results}

\paragraph{Baseline vs self-attentive systems}
We report baseline results in Table~\ref{tab:baselines}.
The best baseline is the TF-IDF model considering
both text and mathematical formulae as input,
it achieves an MRR of 29.9 and an accuracy of 23.8 (dev set).
These results suggest that the task is not trivial,
and that bag-of-words model are insufficiently
expressive to solve it.
In contrast, our best self-attentive model (Table~\ref{tab:conv-results}) 
outperforms all baselines by a wide margin, 
obtaining an MRR of 64.5 and an accuracy of 57.8 (dev set, local decoding).
However, the neural model fails to improve over the baselines in the text-only setting, perhaps
due to the fact that the limitation in this setting is the lack of sufficient information,
which cannot be compensated by a higher model expressiveness.

\paragraph{Global decoding with local training}
In all settings,
the use of global decoding substantially improves accuracy.
This improvement is also manifested with baselines.

\paragraph{Global training} 
We obtain a substantial improvement over local training
when incorporating the global loss.
However, the improvement is much better for models that already have high results (i.e.\ math-only and math-text settings).

\paragraph{Effect of input type}
For baselines, we observe that using both mathematical formulae and text gives the best results.
The baseline models using only text outperform the neural models using the same input as well as the baselines in the math-only settings.
The pattern is different for neural models: the models using only math input are the best and slightly outperform models with both text and math input.
This result suggests that mathematical formulae are crucial to solve the task and best used with an expressive neural model.

\begin{table}
\begin{center}
\resizebox{\columnwidth}{!}{
    \begin{tabular}{l|ccc|c}
    \toprule
    Training & \multicolumn{3}{|c|}{Local} & Global \\

    Decoding & \multicolumn{2}{|c}{Local} & Global & Global \\
    \midrule
    Input   & MRR & Accuracy & Accuracy & Accuracy\\
    \midrule
    Dev & & & \\ 
    \midrule
        Both & 63.2 & 56.1 & 61.4  & 65.6 \\ 
        Text & 21.0 & 15.3 & 16.4  & 18.3 \\ 
        Math & \textbf{64.5} & \textbf{57.8} & \textbf{62.5} & \textbf{67.7} \\ 
    \midrule
    Test & & & \\
    \midrule
        Both & 63.5 & 56.2 & 61.6 & 66.2 \\ 
        Text & 21.6 & 15.8 & 16.6 & 18.1 \\ 
        Math & 64.4 & 57.7 & 62.8 & 67.8 \\ 
    \bottomrule
    \end{tabular}
}
\end{center}
\caption{Self-attentive model results for each setting:
local or global training, local or global decoding.}
\label{tab:conv-results}
\end{table}

\paragraph{Qualitative analysis}
Upon inspection of our global model's incorrect predictions (`both' setting) on the development set, we found that a common source of confusion is due to the proof often introducing discourse-new concepts and new variables,
while not necessarily repeating discourse-given concepts that occur in the statement.
As a result, the set of variables and concepts in a proof might better match those of another statement. We provide examples of the model's output in Appendix~A (supplementary material).
Finally, incorrectly predicted proofs often contain highly polysemous words (\textit{linearly}, \textit{components}) that also occur in the statement.

\section{Conclusion}

We have introduced a new task focusing on the domain
of mathematical research articles.
The task consists in assigning a proof to a mathematical statement.
We have constructed a dataset made of 184k statement-proof pairs for the task
and assessed its difficulty with two classical bag-of-words baselines.
Finally, we have introduced a global neural model for addressing
the structured prediction problem of maximum
weighted bipartite matching.
The model is based on a self-attentive encoder and a bilinear
similarity function.
Our experiments show that bag-of-words baselines are insufficient to solve the task,
and are outperformed by our proposed model by a wide margin.
We found that decoding is crucial to achieve
high results, and is further enhanced by a global training loss.
Finally, our results show that mathematical formulae are the
most informative source of information for the task but are best taken into account with the self-attentive neural model.

\bibliographystyle{acl_natbib}
\bibliography{main}

\appendix

\section{Output Examples}
\label{sec:appendix}

\definecolor{myred}{RGB}{255, 160, 195}
\definecolor{myblue}{RGB}{86, 201, 255}

\DeclareRobustCommand{\r}[1]{{\sethlcolor{myred}{\hl{#1}}}}
\DeclareRobustCommand{\b}[1]{{\sethlcolor{myblue}\hl{#1}}}
\DeclareRobustCommand{\o}[1]{{\sethlcolor{orange}\hl{#1}}}
\DeclareRobustCommand{\om}[1]{\colorbox{orange}{$\displaystyle #1$}}
\DeclareRobustCommand{\rm}[1]{\colorbox{myred}{$\displaystyle #1$}}
\DeclareRobustCommand{\bm}[1]{\colorbox{myblue}{$\displaystyle #1$}}

We provide examples of incorrect outputs by the globally trained model (`both' setting)
in Figures~\ref{fig:example-1} and~\ref{fig:example-2}.
In both cases, the predicted proof contains variable names or concepts from the statement that do not occur in the gold proof.

\begin{figure*}
  \fbox{
  \parbox{\textwidth}{
    \textbf{Statement} (\url{https://arxiv.org/pdf/math/0511162.pdf})\\
    \textbf{Corollary 6.13.} \textit{If \o{$G$} is a \b{compact} \r{connected} \b{Lie group}, then for
    any maximal abelian \r{connected closed} \o{subgroup} of $H<$\o{$G$}, \o{$G$} is the union
    of the conjugates of $H$.}
    \\
    \textbf{Gold proof}\\
    \textit{Proof.} The only thing to observe is that we do not need any
    definability assumptions. The definability comes for free since
    any \b{compact Lie group} \o{$G$} is isomorphic to a compact \o{subgroup} $K$
    of $GL(n,\mathbb{R})$ for some $n$ (see [4, Ch. 3, Thm. 4.1, p.
    136]) and any such $K$ is a (real)algebraic \o{subgroup} of
    $GL(n,\mathbb{R})$ [5, Prop. 2, p. 230], hence it is definable
    in the o-minimal structure $(\mathbb R,<,+,\cdot)$. \qed
    \\
    \textbf{Predicted proof} (\url{https://arxiv.org/pdf/math/0611764.pdf})\\
    \textit{Proof.} By passing to the universal covering of $G$,
    we may assume that \o{$G$} is simply \r{connected}. Theorem 3.18.12 in
    [12] states that in this case every analytic \o{subgroup} of \o{$G$}
    is \r{closed} and simply \r{connected}. Then the result follows from the
    proof of Theorem~3. \qed
    }
  }
  \caption{Example of a wrong prediction, word overlaps are highlighted in orange (present in both gold and predicted proof, red (only in predicted proof), blue (only in gold proof).}
  \label{fig:example-1}
\end{figure*}

\begin{figure*}
  \fbox{
  \parbox{\textwidth}{
    \textbf{Statement} (\url{https://arxiv.org/pdf/math/9902050.pdf})\\
    \textbf{3.10 Lemma} \textit{ Let \b{$G^\circ$} be the identity component of \o{$G$}, let \r{$H$} be a closed,
    connected \r{subgroup} of \o{$G$}, and let $\Gamma$ be a discrete \r{subgroup} of \o{$G$}.
    Then:
     \begin{enumerate}
     \item \label{Gconn-proper}
     $\Gamma$ \r{acts} properly \r{on} $\om{G}/\rm{H}$ if and only if $\Gamma \cap \bm{G^\circ}$
    \r{acts} properly on $\bm{G^\circ}/\rm{H}$.
     \item \label{Gconn-cpct}
     $\Gamma \backslash \om{G} / \rm{H}$ is compact if and only if  $(\Gamma \cap
    \bm{G^\circ}) \backslash \bm{G^\circ} / \rm{H}$ is compact.
     \end{enumerate}
    }
    
    \textbf{Gold proof}\\
    \textit{Proof.} 
    (1) Because every element
    of the Weyl group of \o{$G$} has a representative in \b{$G^\circ$}
    [BT1, Cor. 14.6], we see that \o{$G$} and \b{$G^\circ$} have the same
    positive Weyl chamber~$A^+$, and the Cartan projection $\bm{G^\circ} \to A^+$
    is the restriction of the Cartan projection $\om{G} \to A^+$. Thus, the
    desired conclusion is immediate from Corollary~3.9.
    (2) This is an easy consequence of the fact that $\om{G}/\bm{G^\circ}$ is
    finite [Mo2, Appendix].
    \qed
    \\
    \textbf{Predicted proof} (\url{https://arxiv.org/pdf/math/0209275.pdf})\\
    \textit{Proof.} Because \o{$G$} is reductive, there is a \r{subgroup $H$} of \o{$G$} which is semi-simple
    and such that the quotient $\om{G} \rm{/H}$ is an extension of a finite group by a torus.
    Note that the quotient group $\om{G} \rm{/H}$ \r{acts on} the ring of invariants  $S^H$ for 
    the semi-simple group: $\bar g \in \om{G}\rm{/H}$ \r{acts} on $f \in S^H$ by
    $g\cdot h$ where $g$ is any lifting to \o{$G$} of $\bar g$.
    It is easy to verify that $S^G = {(S^H)}^{G/H}$.
    Because \r{$H$} is semi-simple, the ring $S^H$ is Gorenstein. Thus by
    the preceding lemma, it is strongly F-regular. On the other hand, $\om{G}\rm{/H}$ is 
    linearly reductive and thus
    the inclusion ${(S^H)}^{G/H} \hookrightarrow S^H$ is split by the Reynolds
    operator. This splitting
    descends to characteristic $p$ for all $p>0$. Therefore,
    because $S^H$ is strongly F-regular in almost all fibers,
    so is its direct summand
    $S^G =  {(S^H)}^{G/H}$.
    \qed
    }
  }
  \caption{Example of a wrong prediction, word overlaps are highlighted in orange (present in both gold and predicted proof, red (only in predicted proof), blue (only in gold proof).}
  \label{fig:example-2}
\end{figure*}

\end{document}